Integrating Risk in Humanoid Robot Control for Applications in the Nuclear Industry – 18230

Xianchao Long, Philip Long, Aykut Onol, Taskin Padir*

*RIVeR Lab, Department of Electrical and Computer Engineering, Northeastern University, Boston, MA.

**ABSTRACT**

This paper discuss the integration of risk into a robot control framework for decommissioning applications in the nuclear industry. Our overall objective is to allow the robot to evaluate a risk associated with several methods of completing the same task by combining a set of action sequences. If the environment is known and in the absence of sensing errors each set of actions would successfully complete the task. In this paper, instead of attempting to model the errors associated with each sensing system in order to compute an exact solution, a set of solutions are obtained along with a prescribed risk index. The risk associated with each set of actions can then be compared to possible payoffs or rewards, for instance task completion time or power consumption. This information is then sent to a high level decision planner, for instance a human teleoperator, who can then make a more informed decision regarding the robots actions. In order to illustrate the concept, we introduce three specific risk measures, namely, the collision risk and the risk of toppling and failure risk associated with grasping an object. We demonstrate the results from this foundational study of risk-aware compositional robot autonomy in simulation using NASA's Valkyrie humanoid robot, and the grasping simulator HAPTIX.

**INTRODUCTION**

As nuclear facilities reach the end of their life cycle they must be decommissioned in a safe and efficient manner. During the decommissioning task, the radioactive material is airborne creating an extermly hazardous environment. A typically decommissioning task requires transporting all debris and objects from the interior of the glovebox to the exit port, a dull and repetitive task. Although a specialized robotic system for glovebox operations may be the optimal solution, humanoid robots are an attractive option since they can operate in a variety of environments and use tools that are designed for humans. Classical industrial robots are typically controlled by precisely following a desired position trajectory. The trajectory is generated off-line by an operator or a CAD model. Therefore this method assumes that the environment is static and precisely known and that the robot's location with respect to a fixed frame has been measured during an off-line calibration phase. As a consequence, these systems are installed in highly controlled and isolated environments, where a change in the environment requires a redefinition of the robot's task. The primary motivation behind the use of humanoid robots is to extend the domain of robot applications to unexplored zones with dynamic attributes, thus such off-line planning methods are no longer feasible. In fact to achieve this, it is imperative to equip the humanoid with a vast array of sensing capacities. Though sensors are constantly improving, no apparatus is perfect and frequently they require painstaking calibration. In addition to this, sensors suffer from error propagation over time, notably state estimation drift. Within an industrial process slight degradations in performance can be measured and thus corrected before failures occur. However in safety critical one-off tasks, for instance, handling high-consequence materials in nuclear facilities, it is necessary that the system achieves its goal or at the very least fails in a safe way. With this in mind, we believe it is important that the risks associated with a robot's motion be calculated before execution and that these risks take into account sensors uncertainties, robot state uncertainties and severity associated with failure.

In this paper, we discuss our ongoing work that integrates risk into the robot control architecture. We present two case studies. Firstly, the risk associated with grasping an object with either an uncertain pose or that possibly displace before grasping is complete. In this case, we show that the grasp quality can be affected by this error which in turn increases the chances of failure. Secondly, we propose a risk aware decision making framework for humanoid robots that allows the execution of task-level autonomous behaviors, such as reaching at a specified point in space with either of its arms, grasping, moving 1-step forward, and taking multiple steps for a goal posture. The robot's high degree of freedom allows an infinity of solutions to accomplish a desired task. Instead of selecting the solution that minimizes for instance energy consumption or other performance metrics like classical redundancy resolution schemes, we propose to integrate these metrics with sensors noise covariance in order to obtain a risk level for each solution. The risk associated with each solution is then compared with the solution "payoff".

**Risk in robotics**

Humans often incorporate risk in their decisions, for example in deciding to purchase insurance, cross a busy road, or carrying an uncomfortable large load instead of making several trips. In each case the decision can be viewed in terms of a risk versus reward model and thus the following three components are present: Reward in the case of success, likelihood that the failure will occur and the severity of consequence should failure occur. The risk level associated with an action can been seen as the product of its probability of occurrence times the severity of the consequence as shown in equation (1):

$$Risk = Probability\ of\ accident \times Expected\ loss \in case\ of\ the\ accident \qquad (1)$$

While the probability of an accident, may be calculable within bounds of probability, the expected loss is a metric is subjective. Clearly the expected loss, or cost of behavior [1] is inverse to the expected payoff in the case of success but this is also a somewhat uncertain measure. Take as an example, the St. Petersburg paradox [2] which consists in tossing a coin repeatedly. The jackpot is doubled, each time heads appears, however the game ends once tails appears. The question is how much should the gambler pay to enter the game?  Mathematically, the expected payoff in this case is infinity [3], which is clearly not a very realistic model. Though the paradox remains unresolved, Bernoulli, whose cousin posed the problem, proposed a model based on the relative wealth of the participants i.e. a subjective evaluation of risk based on the player [4]. Thus there is an interpretive/qualitative factor to risk evaluation, which is robotics has led to two main approaches, firstly the treatment of risk as purely a probability and secondly using subjective measures.

Risk can be treated as a probability if we assume that severity is constant, in this case the expected loss in case of the accident is equal to one, i.e. all failures are equally bad. This approach is often used in collision avoidance strategies for mobile robots and autonomous vehicles [5], [6] or in optimum control strategies [1]. However intuitively this approach is incorrect, for instance a head-on collision is infinitely more dangerous than two vehicles scraping up against each other. An alternative idea is to use a known safety criteria to evaluate the severity of possible failures which is especially applicable to the installation of manipulator in an industrial environment. For instance in [7], the authors purpose a robot control scheme that evaluates the severity of a potential collision to changes its controller. In this case an industrial robot cell is monitored. If an operator enters a zone where the collision is potentially dangerous, the robot either slows or stops. The severity is obtained from an empirical data set that measures the danger of collisions with industrial robots [8] a metric that itself is quite quantifiable [9] . Likewise in [10], in a robot assisted needle positioning system, the severity of damage is rated on a scale as is the probability of occurrence. Using both these measures risk minimizations steps can be taken. In both cases, it can be seen that the severity is a global measure that has been empirically obtained before any test has taken place. In our target application, that of manipulating nuclear material in unknown dynamic environment, it is unreasonable, perhaps even unethical, to believe the robot can/should compute severity indexes for everything that can go wrong. Instead, the likelihood of this occurrence should be communicated to a teleoperator who then can make an informed decision, based on their past experiences dealing with hazardous material.

**DESIGN OF A RISK METRIC FOR GRASPING**

Robot grasping is a complex and ongoing research problem. In order to evaluate the success of the grasp, the geometrical relation between the contact points and the object shape must be studied. The relation is known in robotics as the grasp matrix, which relates the twist at the finger frame, denoted $v$ to the twists at the object frame, denoted $v_{obj}$ given as: $v_{obj} = G^T v$ . $G$ is a 6 x (6xn) matrix where n is the number of fingers grasping the object. Intuitively the matrix can be understood by considering the case, where the rank is less than six. In this case there is one dimension of the object twist (i.e. on translational or angular velocity) that cannot be produced by the fingers i.e. the object can move freely in this direction in spite of the current grasp. The ideal grasping situation is known as *form closure*. This means that the object is rigidly fixed independent of the force applied by the grasp. In practice this is difficult to achieve for all but very contrived object forms. A more reasonable metric is known as *force closure*, which essentially means the grasp force can resist the object wrench. Nevertheless, in spite of the simplifications force closure is still a complex problem which depends on the torque capacities of the hand, friction coefficients and the object's inertial properties, many of which cannot be known a priori. Furthermore, often the exact configuration of the object is uncertain meaning the grasp may be executed from an erroneous start position, for example when the object is in motion.

In this study [11], we developed a risk metric associated with the timing of the grasp. For instance, consider the case where the object is moving with translational velocity along the x-axis, should the form of the object vary along this axis, the resulting grasp type is dependent on the execution time. To model this we varied the starting pose of the hand and executed a pre-defined grasp type. Thus unknown motion along the positive x-axis can be modeled as error in the negative x location of the hand. A standard grasp metric Q related to the condition number of $G$ is evaluated after initial grasp and then reevaluated after vigorously shaking the object using a sinusoidal excitation. The simulation environment is shown in Fig.1. A risk metric is defined for both the position and the orientation error of the grasping hand, as shown in equation (2)

$$P_x = \frac{1}{1 + e^{-a\frac{x}{x_{max} - d} + b}} \quad (2)$$

Where *x* is the distance in the translational case and an angle in the revolute case, *a* and *b* are constants. The individual risk terms are combined using a weighted sum. Thus the objective is to investigate if a suitable combination of the positional and angular risk metrics can predict failure for the grasping task. 34 simulations are carried out, the grasp quality is compared before and after the excitation. The resulting positional risk index is shown in Fig. 2 . This figure shows the grasp quality versus the deviation of the pre-grasp position of the hand from the ideal pre-grasp position, which yields the following observations. Firstly, it can be deemed that the grasp quality has a correlation with the pre-grasp pose of the hand since the alteration of the grasp metric across the surface is smooth. Secondly, the grasp quality deteriorates as the hand moves towards right-hand side due to the morphology of an anthropomorphic right hand, namely, the thumb being in the left-hand side of the hand. Finally, it is observed that the grasps with lower height are better, and the best grasp quality value is obtained for the grasp with the lowest height in the left-hand side of the surface.

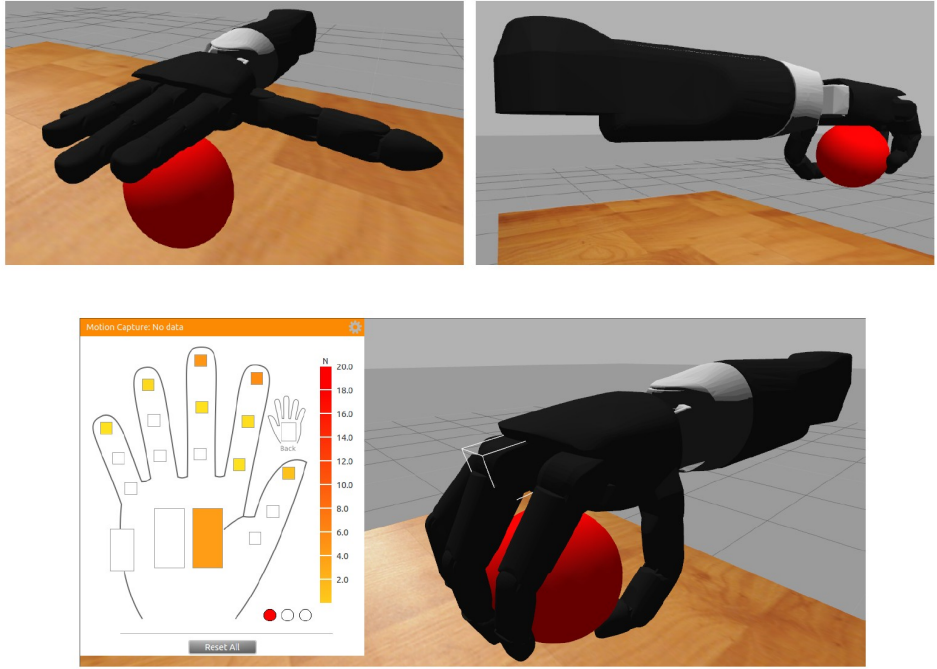

Fig.1: Grasping HAPTIX simulation software

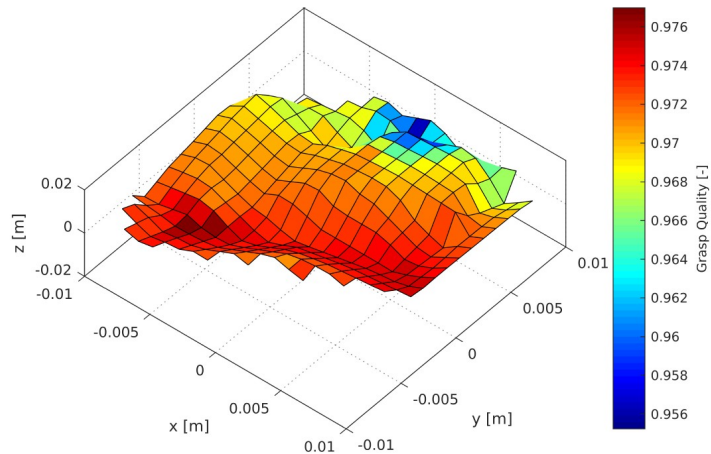

Fig. 2 Variation of the grasp quality with respect to the deviation of the pre-grasp position of the hand.

**DESIGN OF RISK BASED DECISION PROCESS**

Our second study [12] concerns a pick and place operation from a tabletop using a humanoid robot, in this case the NASA Valkyrie robot as shown in Fig. 3. It is assumed that the task can be completed by composing a sequence of behaviors selected from the feasible actions such as picking with the left arm, picking with the right arm, passing the object between hands etc.. The actions make up the task-level robot behaviors and the high-level mission planner (At the DRC, DARPA ROBOTICS CHALLENGE, Finals, the mission planners were operators who had intermittent communication with the robot in the simulated disaster scene.) can achieve compositional robot autonomy by stitching these behaviors together, for example P**ick left** *then* **Place left** or **Pick left** *then* **Hand to right** *then* **Place right** or **Pick right** *then* **Place right.** We consider two specific risk measures for compositional humanoid robot autonomy. More specifically, we will discuss our methodology to evaluate the risk associated with an action composition for completing a given task by taking into account the collision risk and the fall risk.

**Collision Risk**

Several factors could potentially lead to undesired collisions between a robot and obstacles in spite of a predicted collision free motion trajectory. These include observation errors caused by uncertainty in robot's sensor measurements such as cameras, joint encoders and inertial measurement units, and controller errors caused by actuator uncertainties during trajectory tracking. The minimum distance between any point on the robot and any point in the environment is calculated, and denoted as *d*. Provided the minimum distance is less than a reasonable safety margin, the instantaneous risk is then calculated as

$$\left(1 - \frac{d}{d_{saftey}}\right)^b \quad .$$

A trajectory could be ranked according to its maximum risk value, however this could be misleading. For instance is a trajectory that is briefly very close to an object "riskier" than a trajectory than is slightly further away but remains in the proximity for a significant duration?

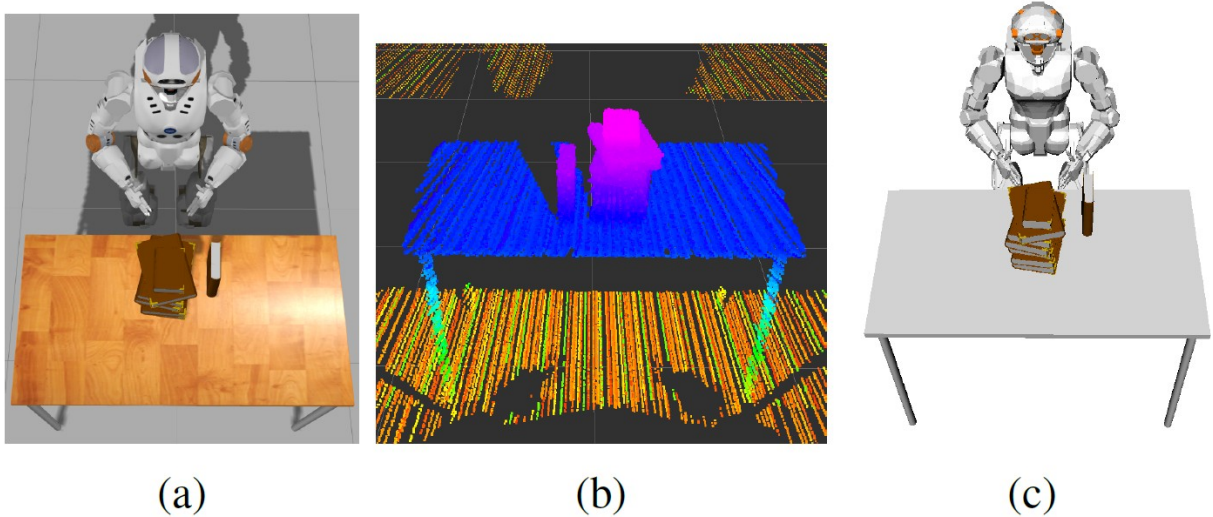

(a)          (b)          (c)

Fig. 3: Book pick-and-place experiment. (a) Experimental setup for the simple pick-and-place scenario in Gazebo. (b) Point cloud data collected by the robot's vision system. (c) A view of the environment in the motion planner.

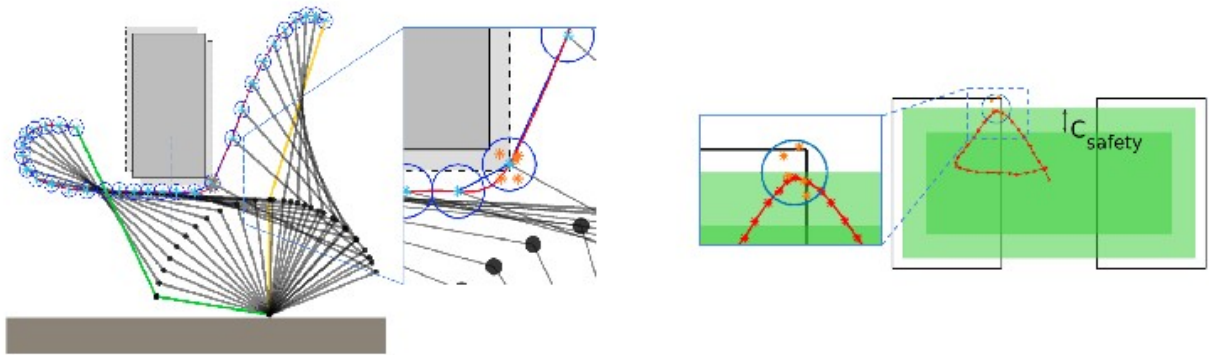

Fig. 4 (right) Illustration of collision risk due to object uncertainty (left) Illustration of fall risk due to support polygon uncertainty

In this case, we assume the risk of collision is related to the both the absolute minimum distance between the robot and all obstacles in the workspace for the entire trajectory and the length of time the robot spends in proximity to the object. The former can be viewed as the risk of collision due to the object being closer reality than in the reconstructed environment. The latter can be view as the risk associated with robot state errors and trajectory deviations. After which the weighted sum is taken to find a true collision risk metric for the entire motion.

**Fall Risk**

Balance is one of the most important criteria for humanoid robots. To avoid toppling over, the robot must keep the vertical projection of its center of mass (CoM) in the support polygon. If the CoM gets close to the boundaries of the support polygon, the risk of failure increases. During the desired motion execution, the state estimator calculates the CoM position from sensor data and the controller attempts to minimize the error between the desired and estimated CoM trajectory. Due to system and measurement uncertainties, the state estimator will have errors in the CoM position calculation resulting in a region of error represented by the blue circle in Fig. 4. If the estimated CoM position moves closer to the support polygon boundary a part of the circular region will fall outside of the support polygon, which means the actual CoM could also fall outside resulting in the most common mode of failure for humanoid robots, falling over. Thus to evaluate the toppling risk, the distance from the projected CoM to the edge of the support polygon is calculated throughout the trajectory.

**Simulation Experiments**

We performed simulation experiments to complete a pick and-place task using NASA Valkyrie humanoid robot. The object to be manipulated is a book as shown in Fig. 3. Our goal is to close the loop to demonstrate the use of total risk measure developed here to enable a complex humanoid robot, Valkyrie, to select risk-averse motion plans by minimizing risks of falling and collisions with the environment. The experiment was implemented in the Gazebo simulation environment. The task involves picking up the book from one side of the stack and placing it on the other side. Assuming known models for the books and the table, the motion planner reconstructs the experiment scenario in its planning environment. Three different action sequences were proposed, picking with the left hand then placing with the left, picking with the right hand and placing with the right hand, finally picking with the left hand, handing the book over to the right hand and placing with the right hand. In the first two cases, three connecting motion trajectories are required while for the third case a fourth motion trajectory is required to hand the book from one arm to another.

The motions corresponding to the three task sequences are planned by the optimization-based motion planner used in the two-link arm, shown in Fig. 4, to generate a sequence of collision free robot configurations.

Fig . 6 shows the minimum distance between the robot and the obstacles along the trajectories for the three compositions, left arm, right arm and dual arm motions. The shortest distance data was separated into four parts based on the different types of motion. In the placing motion part, when using single arm compositions, the robot's motion trajectories are closer to the table and the book stack than the motion trajectory in the dual arm composition. After picking the book, the robot moves around the book stack while holding the book and keeps itself as close as possible to the obstacles in the single arm compositions. However in the dual arm composition, due to the added handover motion, the robot first pulls the book away from the obstacles and delivers it to the other hand.

Fig. 5 shows the COM for the three task sequences. The CoM travels a larger distance and occasionally approaches the edge of the support polygon when the robot uses single arm motions (only using left arm and only using right arm). When following the left arm composition's planed trajectories, the robot shifts its CoM to the upper-right area of the support polygon as it tries to place the book, and maintains the CoM near the support polygon edge in order to avoid the obstacles. Although the CoM trajectory in the right arm composition is closest to the edge of the support polygon, the robot can quickly shift its CoM back to a safer position after it picks and places the book. The robot is able to maintain its CoM around the origin of the support polygon during the dual arm composition. The simulation is repeated ten times, while the position of the book and obstacles were varied, in order to verify if the risk measure is an accurate. The results are shown in Table 1. It is observed that when the robot uses dual arm composition, it can finish the tasks 10 times without falling and with no collisions. In 50% of the tests for the single arm compositions collisions occurred.

TABLE 1: Experimental Results

|  | Left Pick, Left Place | Right Pick, Right Place | Dual arm motion |
|---|---|---|---|
| Collision Risk | 5.9 | 5.4 | 3.0 |
| Fall Risk | 7.9 | 7.0 | 5.0 |
| % Failure Collision | 40% | 20% | 0% |
| % Failure Fall | 20% | 30% | 0% |

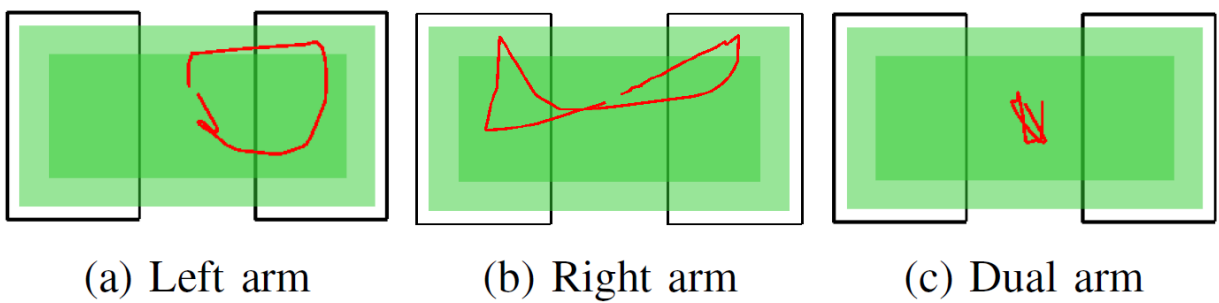

(a) Left arm      (b) Right arm      (c) Dual arm

Fig. 5: Center of mass trajectories as projected on support polygon

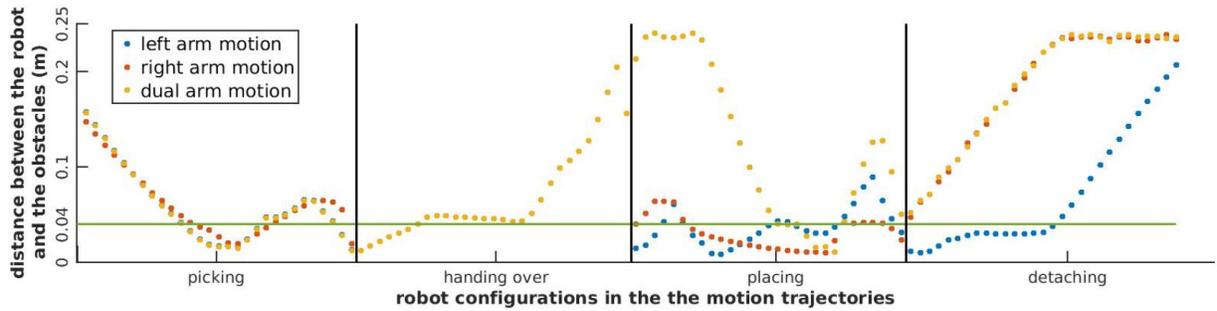

Fig 6: Distance between the robot and the obstacles for the three pick-and-place compositions of motions.

**CONCLUSION**

In this paper, we have a presented an overview of our ongoing work that aims to integrate risk into the robot control architecture. Firstly, we have presented the state of the art regarding the use of risk in robotics. Secondly, we have presented two case studies, predicting grasp failures and integrating risk in the decision process for a humanoid robot. We have shown how the relative error between the hand and the target object can be used to predict chances of failure by executing a series of dynamic simulations with an anthropomorphic hand grasping a simple object. We believe that this method can be used to aid the timing of grasp in robotics an essential tool to increase robot autonomy. For a high degree of freedom robot we have known how possible task solutions can be compared using the risk of collision and risk of toppling as a metric. Our proposed risk metrics have been validated in simulation by showing a correlation between them and failure percentage.

Future work regarding the grasp planning will focus on reversing our approach. In doing so instead of trying to validate a risk metric, we will attempt to use machine learning techniques to discover a risk function from simulations. This function can then be used to estimate the changes of success in unknown scenarios notably in experimental trials.

Currently our risk aware motion planner, uses a weighted sum to combine collision risks and toppling risks, giving more significance to toppling risks. This is a simplification of severity which will be addressed in future work. Indeed, severity index for collisions should increase with velocity and should be higher for more sensitive parts of the robot, for example cameras. Likewise, the severity of the robot toppling should be greater when there are objects nearby. Future work will aim to correct this simplification and experimentally verify the results. Finally we aim to integrate our grasping risk metrics within the risk aware motion planner.

**REFERENCES**

[1] T. D. Sanger, "Risk-aware control," *Neural Comput.*, 2014.
[2] M. D. Weiss and U. S. D. of A. E. R. Service, *Conceptual foundations of risk theory*. U.S. Dept. of Agriculture, Economic Research Service, 1987.
[3] D. A. Braun, A. J. Nagengast, and D. M. Wolpert, "Risk-sensitivity in sensorimotor control," *Front. Hum. Neurosci.*, vol. 5, 2011.
[4] D. Bernoulli, "Exposition of a New Theory on the Measurement of Risk," *Econometrica*, vol. 22, no. 1, pp. 23–36, 1954.
[5] S. Noh and K. An, "Risk assessment for automatic lane change maneuvers on highways," in *2017 IEEE International Conference on Robotics and Automation (ICRA)*, 2017, pp. 247–254.

[6] D. Althoff, J. J. Kuffner, D. Wollherr, and M. Buss, "Safety assessment of robot trajectories for navigation in uncertain and dynamic environments," *Auton. Robots*, vol. 32, no. 3, pp. 285–302, 2012.

[7] M. Ragaglia, L. Bascetta, P. Rocco, and A. M. Zanchettin, "Integration of perception, control and injury knowledge for safe human-robot interaction," in *2014 IEEE International Conference on Robotics and Automation (ICRA)*, 2014, pp. 1196–1202.

[8] S. Oberer-Treitz, A. Puzik, and A. Verl, "Measuring the Collision Potential of Industrial Robots," in *ISR 2010 (41st International Symposium on Robotics) and ROBOTIK 2010 (6th German Conference on Robotics)*, 2010, pp. 1–7.

[9] S. Haddadin, A. Albu-Schäffer, and G. Hirzinger, "Safety Evaluation of Physical Human-Robot Interaction via Crash-Testing, Robotics," in *Proceedings RSS 2007*, Atlanta (USA), 2007, pp. 217–224.

[10] M. Nagel, G. Schmidt, G. Schnuetgen, and W. A. Kalender, "Risk management for a robot-assisted needle positioning system for interventional radiology," *Int. Congr. Ser.*, vol. 1268, no. Supplement C, pp. 549–554, Jun. 2004.

[11] A. Ö. Önol and T. Padir, "Towards Autonomous Grasping with Robotic Prosthetic Hands," in *Proceedings of the 10th International Conference on PErvasive Technologies Related to Assistive Environments*, New York, NY, USA, 2017, pp. 385–389.

[12] X. Long, P. Long, and Padir, T., "Compositional Autonomy for Humanoid Robots with Risk-Aware Decision-Making," in *2017 IEEE-RAS 17th International Conference on Humanoid Robots (Humanoids)*, 2017.